\documentclass{bmvc2k}

\title{You Better Look Twice: a new perspective for designing accurate detectors with reduced computations}

\addauthor{Alexandra Dana}{alex.dana@samsung.com}{1}
\addauthor{Maor Shutman}{maor.shutman@samsung.com}{1}
\addauthor{Yotam Perlitz}{perlitz@gmail.com}{1}
\addauthor{Ran Vitek}{ran.vitek@samsung.com}{1}
\addauthor{Tomer Peleg}{tomer.peleg@samsung.com}{1}
\addauthor{Roy J Jevnisek}{roy.jewnisek@samsung.com}{1}

\addinstitution{
	Samsung Israel R\&D Center\\
	Tel Aviv\\
	Israel\\
}

\runninghead{Dana et al.}{BLT-net}

\def\etal{\emph{et al}\bmvaOneDot}

\usepackage{caption}
\usepackage{amsmath}
\usepackage{amssymb}
\usepackage{booktabs}
\usepackage{enumitem}
\usepackage{float}
\usepackage{algorithm}
\usepackage[noend]{algpseudocode}
\usepackage{enumitem}

\begin{document}

\maketitle

\begin{abstract}
General object detectors use powerful backbones that uniformly extract features from images for enabling detection of a vast amount of object types. However, utilization of such backbones in object detection applications developed for specific object types can unnecessarily over-process an extensive amount of background.
In addition, they are agnostic to object scales, thus redundantly process all image regions at the same resolution.
In this work we introduce BLT-net, a new low-computation two-stage object detection architecture designed to process images with a significant amount of background and objects of variate scales. BLT-net reduces computations by separating objects from background using a very lite first-stage. 
BLT-net then efficiently merges obtained proposals to further decrease processed background and then dynamically reduces their resolution to minimize computations. Resulting image proposals are then processed in the second-stage by a highly accurate model.
We demonstrate our architecture on the pedestrian detection problem, where objects are of different sizes, images are of high resolution and object detection is required to run in real-time. We show that our design reduces computations by a factor of $\times$4-$\times$7 on the Citypersons and Caltech datasets with respect to leading pedestrian detectors, on account of a small accuracy degradation. This method can be applied on other object detection applications in scenes with a considerable amount of background and variate object sizes to reduce computations.
\end{abstract}

%-------------------------------------------------------------------------
\section{Introduction}
Object detection is a fundamental computer vision task with an abundant number of applications. General object detection architectures such as one-stage and two-stage detectors are designed to enable detection of a vast amount of object types in an image, without making any prior assumptions regarding objects' overall image coverage and amount of background. In such setups, it is therefore more computations-wise efficient to uniformly extract features once from the entire image. Over the years, these approaches achieved impressive detection accuracy results on general datasets such as PASCAL VOC \cite{pascal2010} and COCO \cite{coco2014}.

Nonetheless, a variety of real-world applications are developed for detecting specific object classes, such as pedestrians and vehicles in autonomous driving \cite{citypersons,caltech,kitti,eurocity}, buildings and roads in satellite images \cite{fMoW2018}, humans and vehicles in aerial images \cite{dota2018}, faces in surveillance, auto-focus and tagging applications \cite{fddb2010,AFW2012,widerface2016} to name a few. In such scenarios, a non-negligible amount of image regions can be potentially referred to as background. These regions, when detected at a low computational cost and discarded from further processing, could significantly reduce the number of computations required for rich features extraction needed for accurate object detection. In addition, low-computation estimation of objects scale can be utilized to significantly reduce the resolution of candidate image regions, thus further decreasing the overall computations of the detector.

In this work we revisit the two-stage architecture to better utilize both object image coverage and objects scale.
To this end, we suggest BLT-net, a two-stage architecture. BLT-net uses a very lite first-stage  architecture to estimate objects location and their scale and a second-stage that processes these image proposals for accurate object detection.
To further decrease computations of the second-stage, we aim to dynamically produce an amount of proposals that is highly similar to the number of relevant objects in the image, with minimal overlap between proposals. In the second-stage, instead of directly extracting rich features for accurate object detection on proposals at full image resolution, we propose downscaling them to the minimum possible resolution to significantly reduce calculations of this stage.

%-------------FIGURE1--------------------------------------------
\begin{figure}[t]
	\begin{center}
		\includegraphics[width=0.6\linewidth]{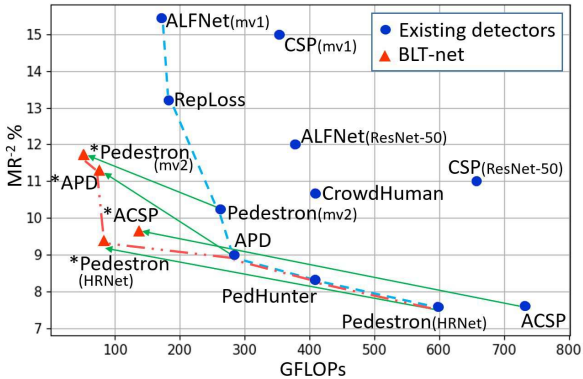}
	\end{center}
	%\vspace*{-5mm}
	\caption{Accuracy (measured by MR$^{-2}$, lower is better)/computations of all pedestrian detectors under 10$^{12}$ FLOPs for the Citypersons reasonable validation dataset (2048$\times$1024 px). The dashed blue line depicts the existing empiric Pareto frontier, green arrows show accuracy/computations change after detectors were integrated into the BLT-net architecture (red triangles), creating the new empiric Pareto frontier (dot-dashed red line).}
	\label{fig:flops_accuracy}
\end{figure}
%---------

For the second-stage one can develop an accurate detector to process image crops. General \cite{cascadercnn,efficientdet,scaledyolo4} and specialized object detectors \cite{najibi2019autofocus,reploss,gao2018dynamic,CSP,uzkent2020efficient,Yang2019ClusteredOD} were shown to achieve high detection accuracy on full-resolution images. We build upon the success of these detectors and apply them directly on our first-stage proposals, with no additional fine-tuning. 

In this work we demonstrate both empirically and analytically the computations advantages of utilizing both image coverage and objects scaling estimation with respect to existing object detection architectures. In addition, this suggested meta-architecture can also be used to reduce computations of existing object detectors that are trained to detect objects in scenes with a significant amount of background, to potentially reduce their computations and thus shortening their inference time \cite{AccSurvey} or enabling them to run on weaker hardware \cite{ignatov2019ai}.

To demonstrate the effectiveness of our approach, in this work we selected to focus on detecting pedestrians in driving scenes, a popular use-case that is highly sensitive to computational load \cite{autolimitations}. This object detection application can benefit tremendously from computational cost reductions due to its requirement to process high-resolution input images for detecting objects at various distances (and thus of various sizes) in real-time \cite{autolimitations}. Our experiments cover the popular Citypersons \cite{citypersons} and Caltech \cite{caltech} datasets. As seen in Figure \ref{fig:flops_accuracy}, our architecture was able to reduce the number of FLOPs\footnote{FLOPs measure mult-add operations \label{ftn:fn1}} by a factor of $\times$4-$\times$7 with a small accuracy (MR$^{-2}$) degradation of 1.4\%-2.3$\%$ of several leading pedestrian detectors.

\noindent In summary, the main contributions of our work are as follows:
\begin{itemize}[noitemsep,topsep=0pt]
	
	\item A new perspective of the two-stage object detection architecture designed to better utilize image object coverage and object scales.
	
	\item A novel method for decreasing computations of existing object detectors for scenes with a significant amount of background with objects of possible variate scales.
	
	\item Theoretical and quantitative analysis of computation reduction aspects of our suggested architecture on the Citypersons and Caltech datasets.
	
\end{itemize}

%------------------------------------------------------------------------
%---------
\section{Related work}

Generally, one-stage or two-stage architectures can be applied to solve various specific object detection use-cases, such as face detection, cars and humans in aerial images, crops in agricultural images, pedestrians and vehicles in self-driving cars. 
To reduce latency and computations and in some cases to increase detection accuracy, over the years architectures were modified or redesigned to better utilize characteristics of the scenery and detected objects, such as foreground/background ratio, objects location and scale distributions.

For example, background region removal for improved computations and latency was demonstrated for pedestrian detection on the Caltech dataset using reinforcement learning techniques \cite{gao2018dynamic,uzkent2020efficient}. In these solutions, the first-step processed downscaled images by a factor of $\times2-\times5$ to detect candidate regions to be processed in the fine-step at their full resolution. The same detection architecture was used in both steps. A high downscaling factor of $\times5$ resulted in a high number of proposed regions that were processed by the second stage to compensate for possible misses. Overall this approach processed $\sim40\%$ of pixels for both downscaling factors. In our work we propose an opposite approach: we process in the first-stage full-resolution images by a very lite detector to enable detection of even small objects at their full image-resolution and then use objects' scaling estimation to reduce image resolution of proposed regions to significantly reduce computations of the accurate detector used in the second-stage, overall achieving a higher computations reduction factor. 

Another approach commonly used for discarding background regions processes images using a cascade of two or more classical \cite{V&J} or CNN based steps \cite{li2015convolutional,yang2016multi,najibi2019autofocus}. This method was successfully applied for detecting faces \cite{li2015convolutional,yang2016multi} or general objects \cite{najibi2019autofocus}. To further reduce computations of later steps, some works united overlapping proposals \cite{chen2016supervised,najibi2019autofocus,Yang2019ClusteredOD}, while others designed their first step architecture to detect regions containing possible clusters of objects \cite{li2020density,Royer2020LocalizingGI}. Nevertheless, these works did not optimize object clustering with respect to their background regions, which can be further reduced to decrease computations. 
In our work we demonstrate that a good scaling estimation of singular objects in the first-stage combined with a cost-efficient area merging criterion can significantly discard additional background areas, thus directly reducing computations of the second-stage detector. In addition, we show that in scenarios where object sizes significantly vary, it is highly beneficial to downscale obtained image proposals, to further decrease computations of the later stage.

In this work we demonstrate our suggested meta-architecture on the computations-sensitive pedestrian detection problem. The majority of pedestrian detectors are based on one-stage or two-stage architectures. One group of works adopted the accurate Faster R-CNN architecture \cite{faster_rcnn} and introduced various adjustments to the common scheme. RepLoss \cite{reploss} altered the training loss, AdaptiveNMS \cite{adaptivenms} refined the NMS step, MGAN \cite{mgan} and PSC-Net \cite{pscnet} added additional branches, PedHunter \cite{pedhunter} introduced a masked guided module for improved detections in crowded scenes,  while Pedestron \cite{pedestron} and CrowdHuman \cite{crowdhuman} trained models on extensive datasets. 
Other works also adopted single-stage architectures developed for general object detection such as SSD \cite{ssd}, YOLO \cite{yolo,yolo2} and others \cite{rfcn,fpn} to potentially decrease inference latency and computations \cite{speed-accuracy}. These include ALFNet \cite{alfnet}, CSP \cite{CSP}, APD \cite{APD} that use various backbones, loss terms, training dastasets and anchor based/free approaches. 
Overall, these detectors use general backbones for feature extraction that result in an increased amount of computations, as seen in Figure \ref{fig:flops_accuracy} and Supplementary Material. In this work we will demonstrate how our proposed architecture enables computation reduction of such existing designs, significantly improving the Pareto frontier.

\section{BLT-net}

\begin{figure}[h]
	\begin{center}
		\includegraphics[width=1.0\linewidth]{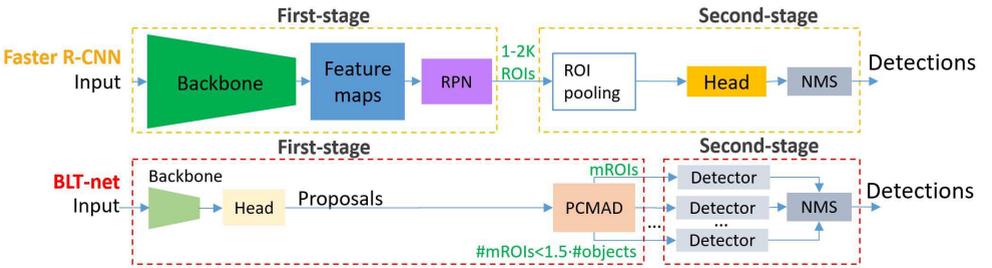}
	\end{center}
	\vspace*{1mm}
	\caption{BLT-net vs. the Faster-RCNN two-stage architecture. The first-stage is replaced by a lite-CNN detector. The ROI pooling module is replaced by the PCMAD algorithm to merge and downscale image proposals. The second-stage head is replaced by an accurate detector that independently processes the merged downscaled proposals (mROIs).}
	\label{fig:overview}
\end{figure}

The BLT-net architecture consists of two stages. The first stage is designed to differentiate objects of interest from the background and estimate their scale using a very lite CNN applied on full-resolution input images. Since objects may overlap each other, same image regions (ROIs) could be unnecessarily processed several times by the second-stage. To this end, overlapping ROIs are merged into mROIs using a cost-effective area criterion. mROIs are then adaptively downscaled using objects' predicted scale to further reduce computations. Downscaled mROIs are fed into the second-stage that consists of an object detector that was trained independently to achieve high detection accuracy.
An overview of the suggested architecture and comparison to Faster R-CNN \cite{faster_rcnn} is described in Figure \ref{fig:overview}.

Let us formulate the number of computations of our approach and analyze the computations reduction factor with respect to an existing architecture integrated into our second-stage. Specifically, let us denote the number of FLOPs per pixel of a general object detector by $A$ and the number of processed pixels in the original image by $N$. Thus the overall computations of a CNN based model are:
\begin{equation}
	FLOPs(OD)=A\cdot N
\end{equation}

\noindent Let us denote the number of FLOPs per pixel of a lite CNN first-stage architecture by $B$ and the total number of pixels in the resulting downscaled mROIs by $M$. Thus the total number of FLOPs in our suggested architecture when incorporating an existing object detector in the second stage is:
\begin{equation}
	FLOPs(BLTnet)=B\cdot N+A\cdot M
\end{equation}

\noindent The FLOPs reduction factor of BLT-net is, therefore:
\begin{equation}
	1/FLOPs_{reduction\_factor}=(B\cdot N+A\cdot M)/(A\cdot N)= B/A+M/N
	\label{eq:1} 
\end{equation}

\noindent  Eq. \ref{eq:1} indicates that the number of computations could be decreased by reducing both the $B/A$ and $M/N$ factors. In this work we suggest a highly efficient first-stage CNN architecture 
to reduce the $B/A$ factor and a novel ROI merging and content-adaptive scaling algorithm to further reduce the $M/N$ factor.

%------------Figure 2 meta-architecture------------------------

%------------------------------------------------
\subsection{First-stage design}
Ideally, we aim at designing a first-stage detector with a high sensitivity and a minimal false-positive rate to reduce the number of processed pixels by the second-stage. More formally, let us define two metrics to evaluate the performance of the first-stage detector: 

1. Sensitivity: the percentage of ground truth bounding boxes in the dataset for which at least $k$\% of their area is covered by mROIs. $k$\% should be set high enough to ensure that adequate image object information is provided to the second-stage for successful detection.

2. Relative processed area ($\overline{M}/N$): the average number of pixels in the dataset processed by the second-stage, relative to the full-resolution image. 

\begin{figure}[h]
	\centering
	\begin{minipage}{2in}
		\includegraphics[width=5.5cm]{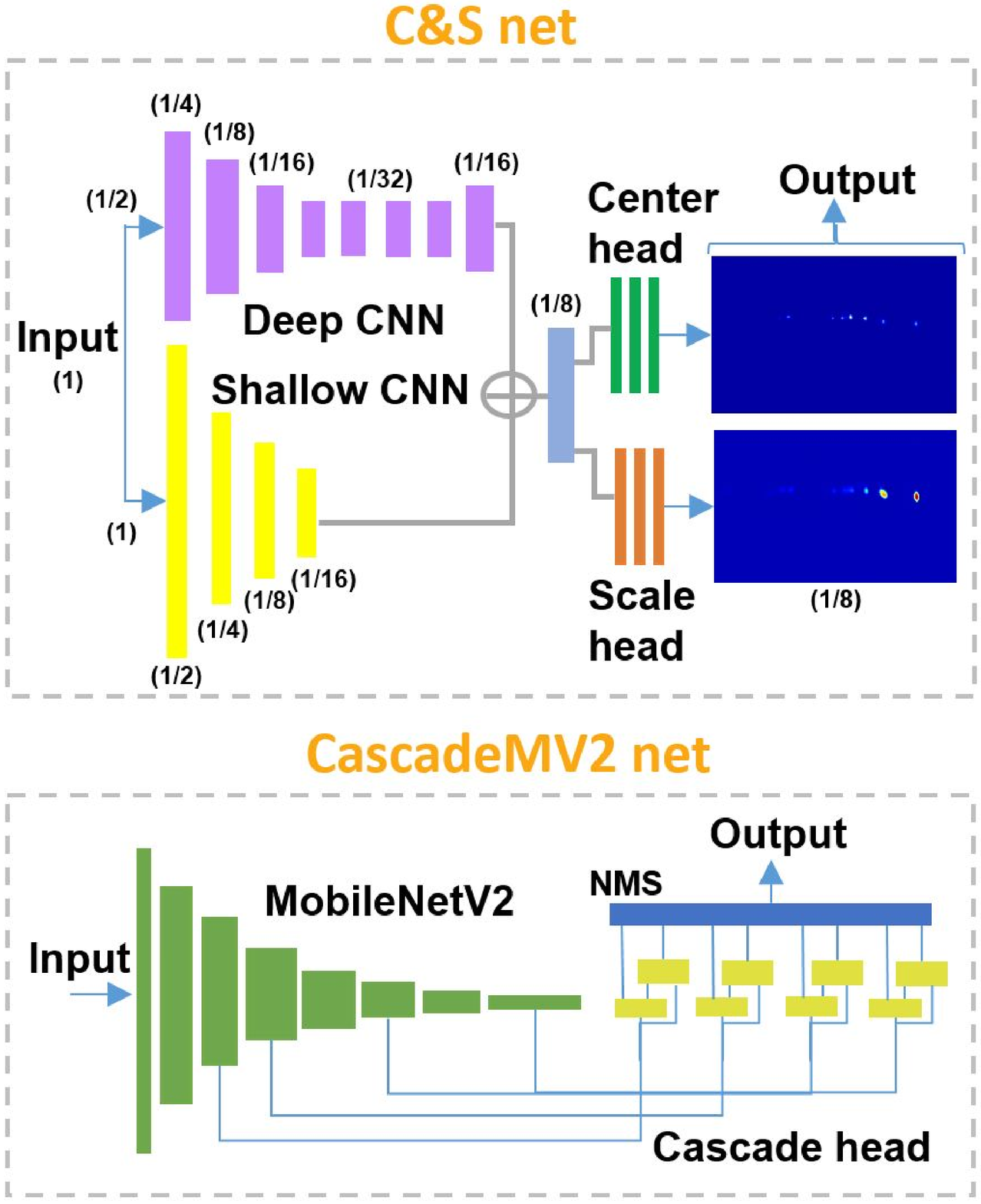} 	
	\end{minipage}
	\hfill
	\begin{minipage}{2.8in}%{0.3\textwidth}
		\algnewcommand{\LineComment}[1]{\State \(\triangleleft\) #1}
		\begin{algorithm}[H]
			\caption{PCMAD}
			\label{alg:pmm_pseudocode}
			\begin{algorithmic}		
				%\small, \footnotesize, \scriptsize, or \tiny
				\scriptsize	
				\State Input: $props_{all}$, $heightRanges$. Init: Empty($mROI_{all}$)
				\For {$hRange$ in $heightRanges$}	 
				\State $props$ = $props_{all} \in hRange$; $Q, mROIs$=\textsc{Init}($props$)
				\While {not empty($Q$)}
				\State	($k$,$j$)=pop($Q$), \textrm{s.t.} min($dist_x$($prop_k$, $prop_j$))
				\If {area($ROI_{kj}$)\textless area($ROI_j$)+area($ROI_k$)}
				\State $ROI_j$=extended(BBox($prop_k$, $prop_j$))
				\State Update pairs in $Q$ with updated $ROI_j$
				\If {cannot pair $ROI_j$ to $ROIs\in Q$} $ROI_j$ $\to$ $mROIs$
				\EndIf
				\EndIf
				\EndWhile
				\State $mROIs \to mROI_{all}$
				\EndFor
				\State \Return $mROI_{all}$
				
				\Procedure {Init}{$props$}
				\State Empty($Q$), Empty($mROIs$)
				\For {each prop $k$,$j$ in $props$}
				\State $ROI_k$,$ROI_j$=extended($prop_k$),extended($prop_j$)
				\State $ROI_{k,j}$=extended(BBox($prop_k$, $prop_j$))
				%\NoThen
				\If {area($ROI_{kj}$)\textless area($ROI_j$)+area($ROI_k$)}
				\State ($k$,$j$) $\to$ $Q$
				\EndIf
				\EndFor
				
				\LineComment{if $prop_k$ is not paired to any other $prop_j$}	
				\For {each $prop_k$ in $props$}	
				\If {($k$, ?) $\notin$ $Q$}  $ROI_k$ $\to$ $mROIs$
				\EndIf 
				\EndFor
				\State Return $Q$, $mROIs$
				\EndProcedure
			\end{algorithmic}
		\end{algorithm}
		
	\end{minipage}
	\vspace{5mm}
	\caption{First-stage components. (a) Two possible lite-CNN architectures used for the first-stage (b) PCMAD pseudo-code.}
	\label{fig:proposal_processing}
\end{figure}

In this work we experimented with two different lite-CNN detection architectures to meet the criteria, one anchor-free and the other anchor-based (see Figure \ref{fig:proposal_processing}). First, we denote the \textbf{anchor-free} architecture by Center \& Scale net (C\&S). The backbone of this architecture consists of two branches that process the image at its original resolution (shallower branch) and at half of the original resolution (deeper branch), a concept previously used \cite{bisenet2018,icnet}. The output resolution of the shallow/deep branches was set to 1/8 and 1/16 respectively. The deep resolution branch output was upsampled to 1/8 of the original image resolution and fused to the output of the shallow branch output.
The network heads output center-likelihood and object-scale maps, similarly to Liu \etal \cite{CSP} . The C\&S net was trained on the Citypersons dataset with crops of size 768$\times$768 px and a batch size of 8. We used random scaling in the range [0.5, 2], random uniform cropping and random mirroring with a probability of 0.5. The model was trained for 32 epochs with a learning rate of $10^{-4}$ and fine-tuned for an additional 350 epochs with a learning rate of $10^{-5}$. While current pedestrian detectors implementations are in the range of 262-867 GFLOPs for detectors with MR$^{-2}$<10\% (see Supplementary Material), C\&S has 27.5 GFLOPs, achieving a maximum ratio $B/A$ of 0.1.

Second, we developed an \textbf{anchor-based} first-stage detector, named CascadeMV2. This detector uses a Mobilenet V2 \cite{mobilenetv2} backbone for feature extraction with a cascade head for improved accuracy \cite{cascadercnn,alfnet}. In this work the detection head was designed using separable convolutions \cite{mobilenets} to reduce computations. Losses were set similarly to anchor-based object detectors \cite{alfnet}. The resulting architecture was trained for 350 epochs on the Citypersons training dataset with a learning rate of $10^{-4}$ and fine-tuned for an additional 100 epochs with a learning rate of $10^{-5}$. For the Caltech dataset, the model trained on the Citypersons dataset was fine-tuned for an additional 100 epochs with a learning rate of $10^{-5}$.
Overall, this detector has 25 GFLOPs for images with a resolution of 2048$\times$1024 px, similarly to the C\&S net. For comparison purposes, the original Cascade R-CNN \cite{cascadercnn} has 262 GFLOPs and ALFNet \cite{alfnet} has 171 GFLOPs for the same image resolution.

Our experiments with designing lite CNNs demonstrated that various architectures could be developed to meet the first-stage criteria, thus we argue that such architecture is not unique and can be constructed using multiple variants or fully automated using NAS methods \cite{nas2019}.

%-------------FIGURE PMM--------------------------------------------

%---------
\subsection{The PCMAD algorithm}

To increase objects' context and compensate for possible localization and scaling errors of the first-stage, object proposals were extended by $p$\% in each direction, creating ROIs. To avoid redundant processing of overlapping ROIs by the second-stage, we developed a \textbf{p}roposals \textbf{c}ontent aware \textbf{m}erging and \textbf{a}dptive \textbf{d}ownscaling algorithm (PCMAD) to produce a set of ROIs with minimal overlapping area. This hierarchical clustering algorithm iteratively merges overlapping proposals with similar heights if the sum of their ROIs area is larger than their union area (united proposals after extension). The proposal pair with the smallest horizontal distance between its centers is selected for merging first. 

The resulting mROIs contain objects of similar heights, all with the original image resolution. However, high-resolution mROIs may be redundant when processing large objects for accurate detection, thus to further reduce computations we conduct mROIs adaptive scaling to a height $h$. This scaling imposes merging object proposals with heights in a specific range only, ensuring that mROIs containing smaller objects are downscaled by a smaller factor than mROIs containing larger objects. In this work height ranges were set to $[h,1.5h)$, $[2h, 2.5h)$, $[2.5h, 3h)$ and $[3.5h, 4h)$.
The $h$ parameter is determined by the second-stage detector minimum resolution requirements for achieving the required detection accuracy. PCMAD complexity is $O(P^{3})$ with respect to the number of proposals $P$.
PCMAD pseudocode is shown in Figure \ref{fig:proposal_processing}.

\subsection{Second-stage architecture}

Generally, the second-stage architecture could be developed from scratch to learn to accurately detect objects from mROIs or one can use existing pre-trained detectors. As current pedestrian detectors achieve highly accurate results on the border of saturation for the Citypersons and Caltech datasets, we opted for integrating such leading pedestrian detectors for the second-stage.
For the Citypersons dataset we integrated Pedestron \cite{pedestron}, APD \cite{APD} and ACSP \cite{ACSP} and for the Caltech dataset we integrated the Pedestron detector. Figure \ref{fig:flops_accuracy} and Supplementary Material show detection accuracy and computations of these detectors. 

The Pedestron detector is based on Faster R-CNN and configured to evaluate during inference 1000 ROIs for images of size 1024$\times$2048 px. Since we apply this detector on smaller image crops (mROIs), we scale down to 64 ROIs per mROI. APD and ACSP are single-stage detectors, thus were applied with their original configuration. 
Final detections from all image crops are processed by a non-maximal-suppression (NMS) algorithm to remove duplicate detections from overlapping mROIs.
The $h$ parameter was set for Pedestron/ACSP/APD to 256/320/384 px accordingly, to conserve as much as possible detection accuracy.

\section{Results}

We start by analyzing the accuracy and computations of all published pedestrian detectors from recent years (see Figure \ref{fig:flops_accuracy} and Supplementary Material). Pedestrian detection accuracy is measured using the MR$^{-2}$ metric \cite{caltech,citypersons} on the reasonable setup and all evaluations were conducted on full images resolution. FLOPs are broadly used today for estimating computations of backbones \cite{mobilenetv2, mv3, shufflenet, efficientdet} and general architectures \cite{tricksbag, lightretinanet}. Although latency and FLOPs are not linearly related \cite{HAT}, previous works \cite{AccSurvey} showed a strong correlation between them. In contrast to latency, which is highly platform dependent, in this work we used FLOPs to allow a direct, platform-independent comparison between various solutions.

\subsection{Evaluating the first-stage architecture}

The C\&S and CascadeMV2 architectures resulted in a B/A factor of 0.046/0.038/0.097 for the Pedestron/ACSP/APD detectors, showing their very lite implementations. Both architectures also achieved a sensitivity of 99\% when setting $k$ to 85\% and $p$ to 10\% (see Table \ref{table:pmm_results}). Moreover, for pedestrians higher than 100 px the C\&S and CascadeMV2 architectures resulted in a sensitivity of 98.26\%/99.47\% respectively, while for pedestrians smaller than 100 px they obtained a sensitivity of 99.16\%/99.08\%, indicating detection robustness to object sizes.
These results confirm that a first-stage architecture is not unique and could be replaced by even lighter designs, as long as they meet the first-stage criteria. Sensitivity could be further increased by selecting proposals with lower confidence levels, on account of processing more proposals in the second-stage, thus enabling to balance between sensitivity and computations reduction of the next stage, without additional training.

Since the lite-CNN architecture is not unique, we also studied a more na\"ive approach that is sometimes used in coarse-to-fine solutions. In these designs an off-the-shelf detection architecture is applied on downscaled input images \cite{gao2018dynamic,uzkent2020efficient}. Similarly, we adopted the single-stage APD detector \cite{APD} since it has the lowest number of FLOPs with MR$^{-2}$ lower than 10\%. To achieve a similar number of FLOPs as our solution, input images were downscaled by a factor of 3.2 on each axis. This resulted in a decreased sensitivity of 72.2\% (see Table \ref{table:pmm_results}), that increased the overall miss rate to 31.9\% (see Table \ref{table:pareto_results}). This accuracy deterioration is expected and impacts most small pedestrians. Specifically, the sensitivity for pedestrians higher than 100 px was 97.34\%, while the sensitivity of pedestrians smaller than 100 px was 35.58\%. Even when downscaling images by a  factor of two on each axis, sensitivity was reduced to 92.1\%, resulting in a miss rate of 15.4\%, showing that using existing detectors for direct inference on downscaled input images is not an effective solution.

\setlength\tabcolsep{4pt} % default value: 6pt
\begin{table}[h]
	\scriptsize
	\begin{tabular}{ll}
		\begin{minipage}{2.6in}
			\begin{tabular}{|@{}llllll@{}|}
				\hline
				Coarse&Data-&PCMAD?&Avg.&Max.&Sensiti-\\ 
				detector&set&&M/N&M/N&vity\%\\
				\hline\hline
				C\&S&CP&x&0.095&1.073&98.9\\
				&CP&\checkmark&0.091&0.38&98.9\\ 
				\hline
				CascadeMV2&CP&x&0.148&1.939&99.2\\
				&CP&\checkmark&0.096&0.56&99.3\\ 
				&CT&x&0.13&1.31&98.58\\
				&CT&\checkmark&0.12&0.95&98.94\\  
				\hline
				APD$\times$1/3.2&CP&\checkmark&0.095&0.55&72.2\\
				APD$\times$1/2&CP&\checkmark&0.12&0.64&92.1\\
				\hline
			\end{tabular}
		\end{minipage}
		&
		\begin{minipage}{3in}
			
			\setlength\tabcolsep{6pt} 
			\renewcommand{\arraystretch}{1.4}
			\begin{tabular}{|@{}llll@{}|}	
				\hline
				Proposals&mROIs&MR$^{-2}$&Avg.\\ 
				extension $p$&downscaled?&&M/N\\
				\hline\hline
				5\%&\checkmark&9.88\%&0.085\\
				10\%&\checkmark&9.39\%&0.091\\
				15\%&\checkmark&9.19\%&0.105\\
				20\%&\checkmark&8.8\%&0.127\\
				\hline
				10\%&x&9.37\%&0.128\\
				\hline
			\end{tabular}
		
		\end{minipage}
	\end{tabular}
	\vspace{2mm}
	\caption{Gauging PCMAD impact. 
		Left: 
		Relative processed area (M/N) and sensitivity of C\&S and CascadeMV2 architectures calculated for $h$=256 px, $k$=85\% and $p$=10\%, evaluated on both Citypersons (CP) and Caltech (CT) datasets. 
		Right: Ablation study on the effect of $p$ and mROIs downscaling on MR$^{-2}$ and M/N, using C\&S in the first-stage and Pedestron (HRNet) in the second-stage on the Citypersons dataset.}
	\label{table:pmm_results}
\end{table}

\subsection{Gauging the PCMAD contribution}
Although produced ROIs achieved high sensitivity, the maximum relative processed area ($M/N$) in some images exceeds 100\% (see Table \ref{table:pmm_results} and Supplementary Material). For $h$=265 px the PCMAD algorithm was able to limit M/N in the worst-case scenario to below 0.56 for the Citypersons dataset, with an average $M/N$ below 9.6\%. Overall, PCMAD significantly reduced the amount of processed pixels by the second-stage detector, contributing to the overall BLT-net computations reduction.
Downscaling mROIs did not decrease detection accuracy (see Table \ref{table:pmm_results}) for the selected $p$ and $h$ parameters, but significantly decreased M/N. Detection accuracy was impacted by $p$, where an increased $p$ improved detection accuracy on account on increasing the M/N factor. Based on this analysis, we select for all the next analyses $p$=10\%, that balances between a low MR$^{-2}$ and a low N/M factor. 

The number of mROIs divided by the number of pedestrians (Citypersons) when using the C\&S / CascadeMV2 was smaller than 1.26/1.31 respectively for the different second-stage detectors. These results indicate that the first-stage was able to optimize the number of mROIs, similarly to the number of ground truth objects, as opposed to the thousands of ROIs used in general two-stage detectors \cite{fast_rcnn,faster_rcnn}.

\subsection{BLT-net impact on computations and detection accuracy}

Overall, BLT-net significantly reduced the computations of the evaluated pedestrian detection architectures by a factor of $\times$4-$\times$7 (see Table \ref{table:pareto_results}). Computations reduction was most significant for Pedestron, which had the highest detection accuracy for small pedestrians, that in turn enabled a more significant mROI downscaling. The total number of FLOPs when using Pedestron or APD in BLT-net second-stage was below 83 GFLOPs, well below all published pedestrian detectors so far. BLT-net increased MR$^{-2}$ by 1.5\%-2.3\% when using C\&S and by 1.5\%-2.64\% when using CascadeMV2 for the Citypersons dataset with respect to the original architectures, with Pedestron being the least impacted detector and APD the most impacted detector. Similar results were obtained for the Caltech dataset.

Although $\sim1\%$ of misses resulted from the first-stage detector, some additional misses resulted from applying the fine pedestrian detectors on mROIs. Such misses are characterized mostly by (a) pedestrians appearing in crowds/occluded or (b) small and poorly illuminated pedestrians. This deterioration is also reflected on the MR$^{-2}$ of the heavy occluded pedestrians, (see Table \ref{table:pareto_results}). All additional misses created by BLT-net's first and second stages on the Citypersons validation dataset can be seen in the Supplementary Material. Accuracy degradation in such scenarios was shown to be improved by extending proposals margins ($p$) on account of increasing second-stage computations, as seen in Table \ref{table:pmm_results}. Such  parameters optimization can be used for balancing between accuracy and computations reduction, without any additional training. Alternatively, detection accuracy could be improved by training second-stage detectors directly on mROIs \cite{Yang2019ClusteredOD,hong2019patch,li2020density} for given $p$ and $h$ parameters.

\setlength\tabcolsep{2.8pt} 
\renewcommand{\arraystretch}{1}
\begin{table}[h]
	\small
	\caption{Detection accuracy and FLOPs of original pedestrian detection architectures and when integrated into BLT-net on full images. Small/large pedestrians are defined to be lower/higher than 100 px. Reasonable/bare/partial/heavy are as defined by the dataset.}
	\vspace{2mm}
	\begin{tabular}{|lllllllllll|}
		
		\hline
		Detector&Dataset&BLT-net&GFLOPs&FLOPs&MR$^{-2}$&MR$^{-2}$&MR$^{-2}$&MR$^{-2}$&MR$^{-2}$&MR$^{-2}$\\ 
		&&first&&reduct.&Reas.&Bare&Partial&Heavy&Large&Small\\
		&&stage&&factor&\%     &\%    &\%       &\%     &\%     &\%  \\
		\hline\hline
		
		Pedestron&CP&-&596.8&-&7.6&5.8&6.3&34.3&5.4&7.6\\
		(HRNet)&CP&C\&S&82.2&7.3&9.4&7.3&8.3&38.5&7&9.5\\
		&CP&CascMv2&82.7&7.2&9.1&6.9&8.2&38.9&6.6&8.8\\
		\hline
		
		Pedestron&CP&-&262.2&-&10.2&7.4&9.1&43.8&7.2&10.1\\
		(mv2)&CP&C\&S&51.5&5.1&11.7&8.2&11.4&48.2&8.1&12.6\\
		&CP&CascMv2&50.2&5.2&11.8&8.2&10.9&47.4&8.6&10.7\\
		
		\hline
		ACSP&CP&-&730.5&-&7.6&4.9&6.9&42.2&4.7&7.2\\
		CP&CP&C\&S&136.7&5.3&9.6&6.3&9.3&47.5&5.9&10.6\\
		&CP&CaMV&128.2&5.7&9.2&5.9&9.0&45.4&5.6&10.1\\
		\hline
		
		APD&CP&-&283.3&-&9&5.2&8.98&47.0&4.5&9.8\\
		&CP&C\&S&75&3.8&11.3&7.8&10.1&51.3&7.8&11.6\\
		&CP&CascMv2&72.7&3.9&11.6&7.3&10.7&50.0&7.4&11.7\\
		&CP&APD$\times$1/3.2&69.5&4.1&31.9&27.7&31.9&55.9&8.2&61.5\\
		&CP&APD$\times$1/2&125.5&2.3&15.4&10.7&15.2&47.2\%&6.8&22.2\\
		
		\hline
		Pedestron&CT&-&87.4&-&2.22&-&-&-&-&-\\
		(HRNet)&CT&CascMv2&13.87&6.3&3.63&-&-&-&-&-\\
		\hline
		
	\end{tabular}
	\label{table:pareto_results}
\end{table}

\section{Conclusions}
In this work we revise the two-stage architecture by leveraging object scales and coverage characteristics to reduce computations. Our suggested design significantly reduced computations of existing pedestrian detectors while achieving similar detection accuracy, creating a new empirical Pareto frontier. This architecture could be applied on other object detection scenarios with similar data characteristics to benefit from significant computations reduction.

\section{Acknowledgments}
We would like to thank Lior Talker for the numerous insightful discussions on the various analyses presented in the paper and Matan Shoef for his contribution to the development of the C\&S net architecture.
%-------------------------------------------------------------------------
\newpage
\nocite{*}

\bibliography{ms}

\end{document}

\maketitle

\section{Detection accuracy and computations of all published pedestrian detectors}
Table \ref{table:benchmark} summarizes detection accuracy and computations measured by FLOPs of the latest published pedestrian detectors evaluated on the Citypersons reasonable validation dataset, on the original image size (2048$\times$1024 px). Lower MR$^{-2}$ indicates higher detection accuracy. 
\begin{table}[h]
	\caption{Pedestrian detectors benchmark on Citypersons validation dataset. Detectors marked in bold are Pareto-efficient according to empirical findings in Figure 1 (main text). For models with published code FLOPs were directly measured, otherwise estimated using the Pedestron framework. The following backbone acronyms were used: MobileNet-V1(MV1); MobileNet-V2(MV2); ResNet-50(RN50);  ResNet-50(RN101).}
	\begin{tabular}{|@{}llll@{}|} %|@{}lllll@{}||lllll|
		\hline
		Detector&Measured?&GFLOPs&MR$^{-2}$\\
		\hline \hline						
		\textbf{ALFNet (MV1)}&\checkmark&171&15.45\%\\
		\textbf{RepLoss (RN50)}&$\times$&183&14.6\%\\
		\textbf{Pedestron (MV2)}&\checkmark&262&10.24\%\\	
		\textbf{APD (DLA34)}&\checkmark&283&9\%\\
		\textbf{PedHunter (RN50)}&$\times$&409&8.32\%\\
		\textbf{Pedestron (HRNet)}&\checkmark&597&7.6\%\\
		
		\hline
		
		CSP (MV1)&\checkmark&352&15\%\\
		ALFNet (RN50)&\checkmark&377&12.01\%\\
		
		CrowdHuman (RN50)&$\times$&409&10.67\%\\
		CSP (RN50)&\checkmark&657&11\%\\
		ACSP (RN101)&\checkmark&731&7.63\%\\
		Citypersons (VGG16)&$\times$&867&14.6\%\\
		AdaptiveNMS (VGG16)&$\times$&867&11.9\%\\
		OR-CNN (VGG16)&$\times$&867&12.8\%\\
		PSC-Net (VGG16)&$\times$&867&10.5\%\\
		MGAN (VGG16)&$\times$&894&11.3\%\\
		One-and-half (RN50)&N/A&N/A&8.12\%\\
		\hline
		%\bottomrule
	\end{tabular}
	\label{table:benchmark}
\end{table}

\newpage
\section{First-stage resulting relative processed area}
\begin{figure}[h]
	\begin{center}
		\includegraphics[width=1.0\linewidth]{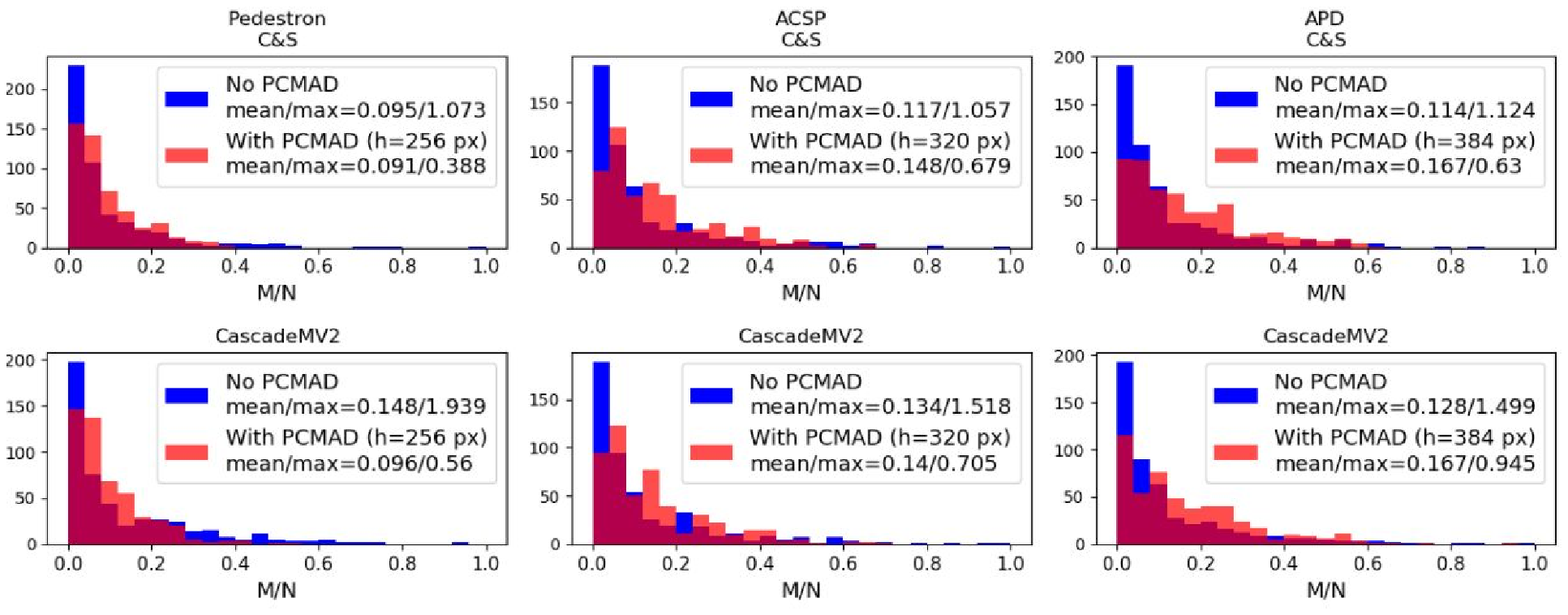}
	\end{center}
	\caption{M/N (relative processed area) distribution per image for the Citypersons validation reasonable dataset, when using C\&S net (top row) or CascadeMV2 net (bottom row) for the various evaluated pedestrian detectors. Red/blue colors depict the N/M distributions with/without applying the PCMAD algorithm. The $h$ parameter indicates the specific heights mROIs were downscaled to before applying the second stage detector.}
	\label{fig:pmm_dist}
	
\end{figure}

%-----------------------------------------------------------------------
\section{BLT-net additional misses}

\begin{figure}[h]
	\begin{center}
		%		\fbox{\rule{0pt}{2in} \rule{.9\linewidth}{0pt}}
	
		\includegraphics[width=1.0\linewidth]{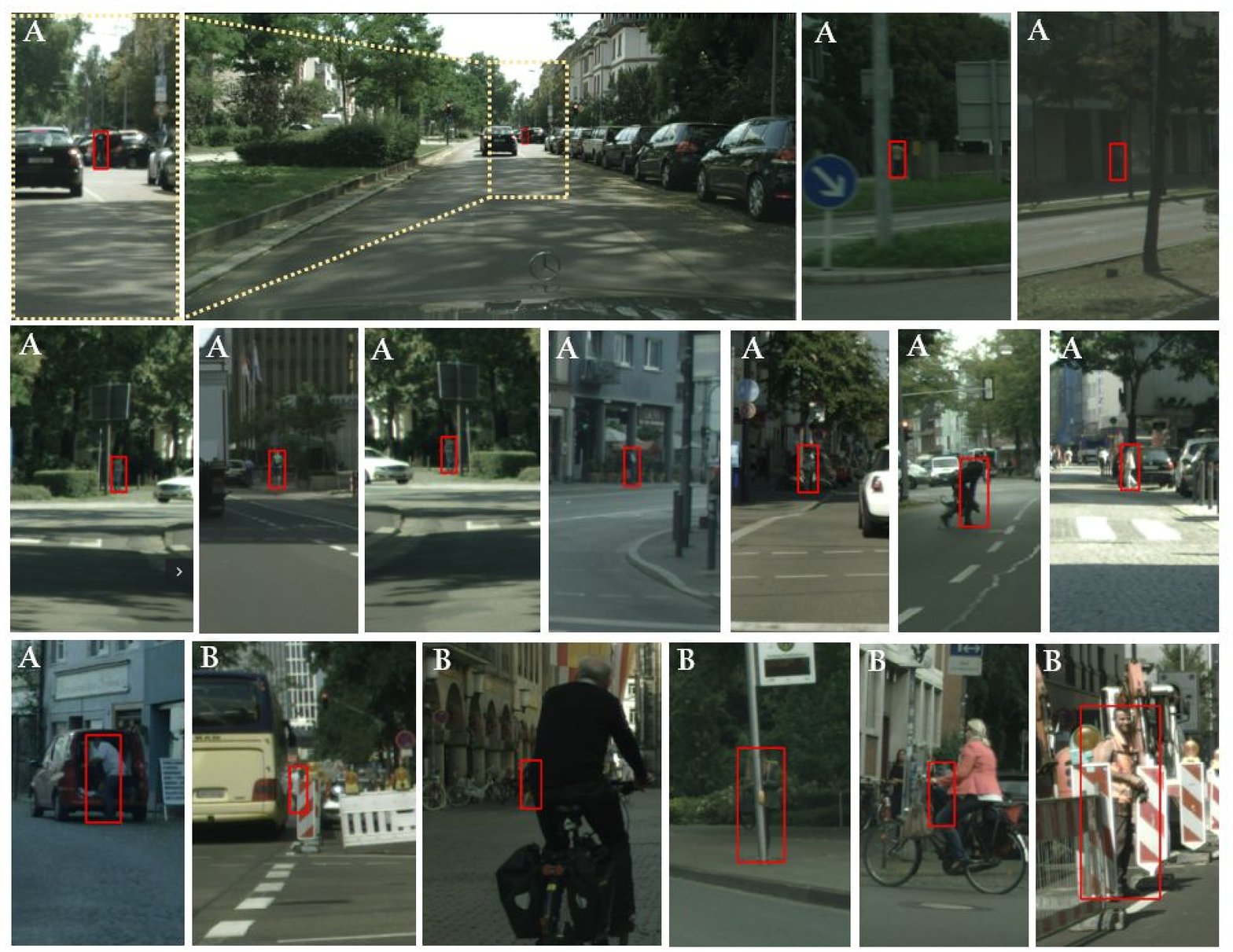}
	\end{center}
	\caption{All first-stage additional misses for the Citypersons validation reasonable dataset, with respect to Pedestron(HRNet) applied on the entire image. The first-stage was implemented using the C\&S net. Small bare and/or poorly illuminated pedestrians are labeled with (A) while partially occluded pedestrians are labeled with (B). All image crops have the same resolution and misses are marked with a red bounding box. The relative crop size with respect to a full image is shown in the top row.}
	\label{fig:all_coarse_misses}
\end{figure}

\begin{figure}[h]
	\begin{center}
		%		\fbox{\rule{0pt}{2in} \rule{.9\linewidth}{0pt}}
		%		\includegraphics[]{overview.png}
		\includegraphics[width=1.0\linewidth]{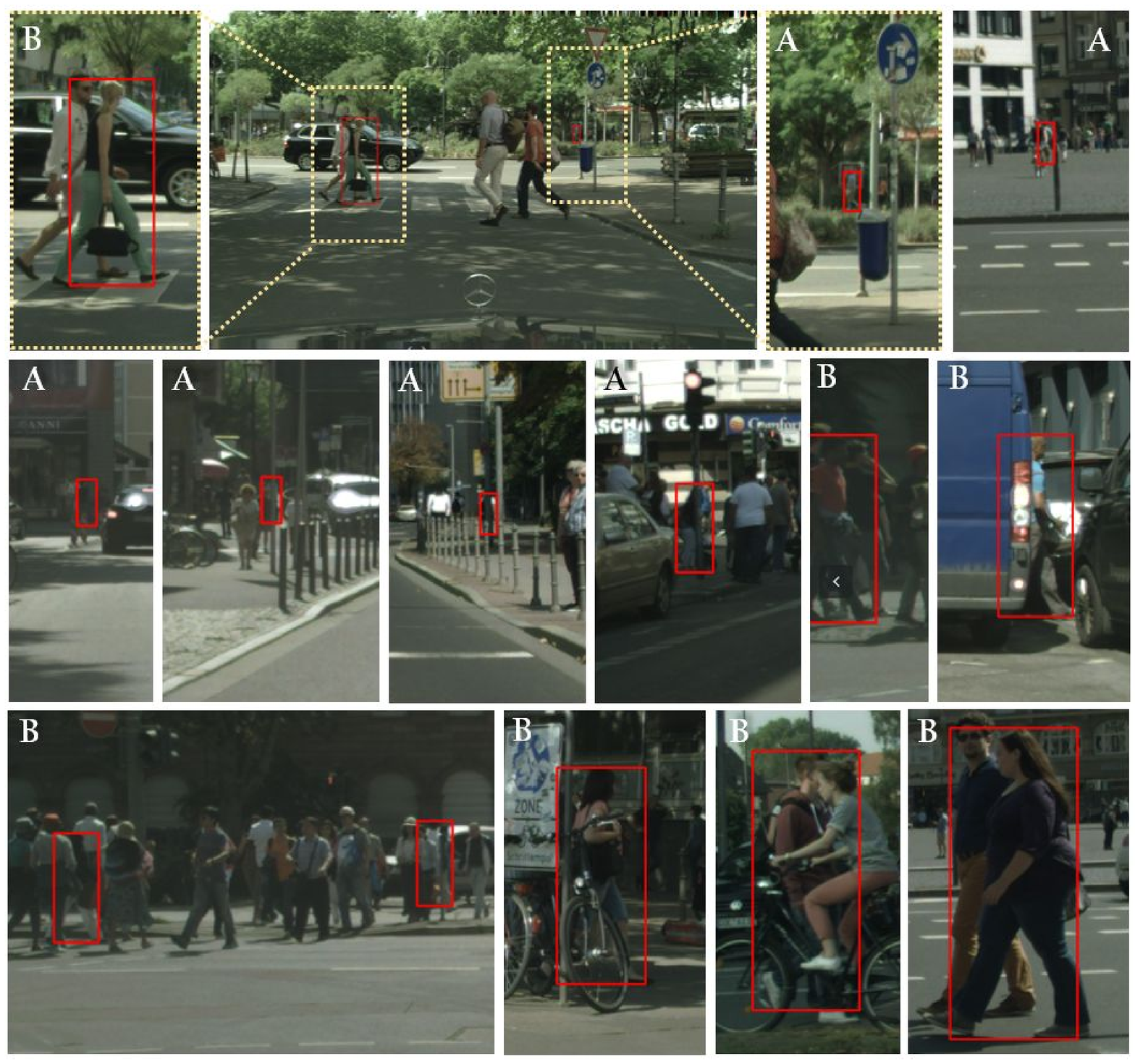}
	\end{center}
	\caption{All second-stage additional misses for the Citypersons validation reasonable dataset, with respect to Pedestron(HRNet) applied on the entire image. The second-stage detector was implemented using the Pedestron(HRNet) detector. Small bare and/or poorly illuminated pedestrians are labeled with (A) while partially occluded pedestrians are labeled with (B). All image crops have the same resolution and misses are marked with a red bounding box. The relative crop size with respect to a full image is shown in the top row.}
	\label{fig:all_fine_misses}
\end{figure}

In this section we show all additional misses of BLT-net when using C\&S net in the first-stage and  Pedestron(HRNet) in the second-stage. Overall, BLT-net achieved a $MR^{-2}$ lower by 1.79\%. All additional misses resulting from the first-stage only are shown in Figure \ref{fig:all_coarse_misses} and all additional misses resulting from the second-stage only are shown in Figure \ref{fig:all_fine_misses}. These results show that misses from both stages are characterized by small and/or poorly illuminated  or occluded pedestrians.

%--------------------------------------

\newpage